# Answering questions by learning to rank - Learning to rank by answering questions

**George-Sebastian Pîrtoacă, Traian Rebedea, Ștefan Rușeți**

University Politehnica of Bucharest, Faculty of Automatic Control and Computers

gpirtoaca@gmail.com, {traian.rebedea, stefan.ruseti}@cs.pub.ro

## Abstract

Answering multiple-choice questions in a setting in which no supporting documents are explicitly provided continues to stand as a core problem in natural language processing. The contribution of this article is two-fold. First, it describes a method which can be used to semantically rank documents extracted from Wikipedia or similar natural language corpora. Second, we propose a model employing the semantic ranking that holds the first place in two of the most popular leaderboards for answering multiple-choice questions: ARC Easy and Challenge. To achieve this, we introduce a self-attention based neural network that latently learns to rank documents by their importance related to a given question, whilst optimizing the objective of predicting the correct answer. These documents are considered relevant contexts for the underlying question. We have published the ranked documents so that they can be used off-the-shelf to improve downstream decision models.

## 1 Introduction

The article at hand devotes to the problem of answering multiple-choice questions where the input consists of an inquiry expressed solely in raw natural language along with a small set of candidate answers (usually 4) from which only one is correct. Moreover, we are targeting questions from a field of science (e.g. chemistry, biology) as they are distinctly more challenging to answer than regular questions (Clark et al., 2018). A relevant example of such a question is the following:

*Which of the following is an example of a physical change? (A) Lighting a match (B) Breaking a glass (C) Burning of gasoline (D) Rusting of iron*

An important characteristic of all solutions developed for this task is that they are not given explicitly any external information in the form of documents supporting the correct answer or semi-structured information. However, external information is highly desirable, especially domain and common-sense knowledge. Thus, most of the state of the art solutions (Pîrtoacă et al., 2018; Nicula et al., 2018; Ni et al., 2018; Zhang et al., 2018) are using a two-step architecture, as shown in Figure 1. In the first phase, an information retrieval (IR) engine *lexically* searches for relevant supporting paragraphs in Wikipedia and other corpora considered relevant for the task. Having extracted some pertinent paragraphs (usually only one per candidate answer), potentially containing relevant information, a machine learning model is employed in the second step to reason about tuples *(question, candidate answer, external information)*. Various downstream decision models are trained to infer whether the candidate answer is correct given the external information: transformers (Pan et al., 2019), attention models (Clark et al., 2018; Ni et al., 2018), or support vector machines (Clark et al., 2016).

However, it has been previously reported that a retrieval-based engine alone is not able to return relevant documents from the reference corpora (Pîrtoacă et al., 2018; Zhang et al., 2018). Depending on the dataset used as an external reference, about 50% of the questions have insufficient and irrelevant support from a standard retrieval model (Zhang et al., 2018). Therefore, the decision engine that relies on the quality of the supporting documents will be highly affected. Irrelevant documents disturb the training process since the models are trained on "wrong" (or noisy) data. As a result, the end-to-end performance of question answering (QA) systems decreases considerably. This article improves the current retrieval scheme by *semantically* ranking the retrieved documents, placing the most relevant ones at the top with a neural self-attention model.

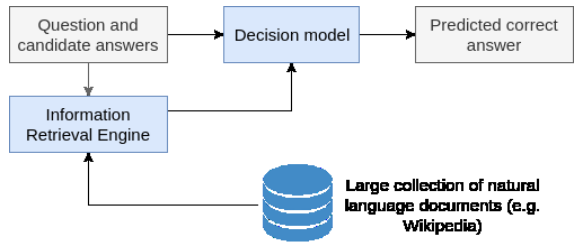

Figure 1: The general high-level architecture of a question answering system

We conduct our experiments on two significant datasets containing science-related multiple-choice questions collected from official examinations: ARC Easy and ARC Challenge (Clark et al., 2018). The former consists of easier questions, whereas the latter incorporates more difficult queries that require inference and external knowledge. In the results section, we are going to observe that our approach achieves state-of-the-art accuracy on both datasets whilst acknowledging their core difference stated by the large gap in the performance: about 28% difference in the accuracy between Easy and Challenge questions.

In this paper, we make the following important contributions:
1) We introduce a neural network architecture, called Attentive Ranker that latently learns to rank supporting documents at a semantic level. The classifier is trained to predict the correct answer to a multiple-choice question achieving state of the art accuracy on the two ARC datasets: Easy (72.30%) and Challenge (44.72%).
2) We show that the set of documents computed by our semantic ranking system can be off-the-shelf adopted by various downstream classifiers to boost their performance (by up to 7%).

## 2 Related Work

A lot of effort has been invested to extract more relevant external documents from natural language corpora to improve QA systems. One approach is to identify the essential terms in questions (Pîrtoacă et al., 2018; Khashabi et al., 2017) and use these terms to improve the quality of the extracted documents. Essential terms are particularly useful for long questions that are intentionally injected with noisy information by teachers to assess the reading comprehension of examined students (which is the case of the ARC dataset, collected indeed from examinations).

Pîrtoacă et al. (2018) propose a neural network architecture, with a small number of parameters, trained to annotate each term in the question with an essentialness score from 0 to 5. The architecture is based on LSTM units (Hochreiter and Schmidhuber, 1997) but with some pre-computed features added to the input to overcome the problem of small datasets: part of speech, a concreteness score for each term (Brysbaert et al., 2014), dependency relations from the parsing tree. The authors show that using the essential terms to extract documents improves the accuracy of multiple-choice QA systems by up to 4% on the ARC dataset (versus a standard IR approach). This highlights that more relevant documents are retrieved using essential terms.

In a similar spirit, Khashabi et al. (2017) report the importance of essential terms for the same multiple-choice QA task but on different datasets. They propose an essential term classifier in the form of a linear Support Vector Machine (SVM) trained on syntactic and semantic features extracted from the question. In total, about 120 types of features (with their combinations) are fed into the classifier. Incorporating the computed essential terms in the IR engine yields an increase in the number of questions correctly answered by up to 5% on the *REGENTS* and the *AI2PUBLIC* datasets[1].

Furthermore, Jansen et al. (2017) formulate the multiple-choice question answering problem into extracting answer justifications and then ranking them. A perceptron is trained to jointly rank the answers along together with their justifications considering the relevance of justifications as a latent variable. The ranking model is tested on a corpus of 1,000 primary school questions, answering 44% of the questions correctly, empirically showing that about 57% of the justifications are meaningful.

A similar approach to the one proposed in this paper (independent research) has been submitted by Banerjee et al. (2019). Their system learns to rank semantically relevant knowledge sentences for a given *(question, answer)* hypothesis. BERT (Devlin et al., 2018) is used as the underlying textual similarity model.

---
[1] Available at http://allenai.org/data.html

On another note, the state of the art downstream decision models are mainly based on transformers (Vaswani et al., 2017) and pre-retrained language models (Devlin et al., 2018; Radford et al., 2019). Sun et al. (2018) present some reading strategies for machine reading comprehension such as back and forth reading or highlighting. These strategies are carried out by fine-tuning a generative transformer (GPT) (Radford, 2018). Their model was the state of the art approach on the ARC dataset (both Easy and Challenge) prior to our proposed model.

Previous work has shown that the performance of models on QA datasets drops dramatically when a document is not provided to support answering the question. Chen et al. (2017) proposed a model, called DrQA, that was trained on the SQuAD 1.1 dataset (Rajpurkar et al., 2016) to find the correct answer to open-domain questions. When a valid document is provided - guaranteed to contain the correct answer - the exact match (EM) score obtained by DrQA is 69.5. However, when the supporting document has to be retrieved from Wikipedia, by an information retrieval engine, the EM score drops to 27.1. This suggests that research has to be invested in improving retrieval strategies and candidate document ranking methods.

As this related work section suggests, the main direction currently approached for increasing the quality of the extracted documents is to employ essential term information in a form or another and blend this knowledge in the query sent to the retrieval engine. Our paper proposes a completely different strategy, which is based on semantically ranking the extracted documents using a neural network that learns to select the most meaningful and discriminative documents for a given question.

## 3 Proposed approach

We briefly describe how an IR engine is used to extract external information (e.g. supporting documents) from natural language resources like Wikipedia to support QA systems. Then, the paper continues with our proposed approach towards rectifying the shortcomings of the current retrieval-based extraction methods.

### 3.1 Extracting supporting documents

It is undeniable that external information is deeply required in some form or another for answering a given question. Humans capture that information by learning, by experiencing and from common-sense knowledge. For machine learning models, external information is injected as input during training. Given a question and a candidate answer, documents containing information relevant to each *(question, candidate answer)* pair are usually extracted from raw text corpora (Pîrtoacă et al., 2018, Nicula et al., 2018). For the example question mentioned earlier in this paper, a good supporting document would be the following one: *"When a glass is broken, a lot of small sharp glass pieces are formed and spread around."* Given this information, a decision component such as a neural network should be able to deduce that "*breaking a glass*" is indeed an "*example of a physical change*", thus predicting the correct answer.

Natural language corpora like Wikipedia, ARC Corpus[2], or a large collection of science books crawled from the World Wide Web (Pîrtoacă et al., 2018) are split into articles, paragraphs, or even at the sentence level and then indexed using various search engines: Lucene (Białecki et al., 2012) or ElasticSearch (Gormley and Tong, 2015), to make the entire query process a lot faster. After the indexing phase is completed, queries are sent to the IR engine in the format *"question tokens" AND "answer tokens"*, thus retrieving documents that contain tokens from the question and tokens from the answer (at least one from each) which are, desirably, semantically relevant – and they should be up to one point, depending on the question difficulty and how it is formulated. Notice that for each candidate answer, a set of documents are retrieved and they will be used by some decision engine to derive if the candidate answer is correct or not. In some cases, no decision engine is used at all and the candidate answer with the highest matching score (for example, TF-IDF (Manning et al., 2010)) is predicted as the correct one. During this paper, we will refer to this simple approach as the "*IR baseline*".

In this paper, we will build upon the aforementioned IR baseline by adding a layer on the top of it, which is a neural network capable of latently learning to better rank the documents (as compared to the IR baseline).

---
[2] http://data.allenai.org/arc/arc-corpus/

## 3.2 Towards a better retrieval engine

An important problem of QA systems is the IR approach for extracting relevant documents. Using a token-based retrieval (Lucene, ElasticSearch) and keeping *the most relevant document* as dictated by a *lexical score* (e.g. TF-IDF, Okapi BM25 (Manning et al., 2010)) does not produce relevant supporting documents for a significant proportion of the questions (Zhang et al., 2018). Our work is a trade-off between the computational performance of such systems and the expressivity of semantic level retrieval. Our main research hypothesis is that semantically ranking the first $N$ (e.g. $N = 100$ or $500$) returned documents instead of sorting them by a lexical metric produces better results and delivers more meaningful documents for answering the question at hand. In other words, some better documents can be found in the first $N$ (where N is a small number), but the best is not always the one with the highest lexical (e.g. TF-IDF) score. We are going to validate our hypothesis in the results section by showing that we can achieve state-of-the-art accuracy on both ARC Easy and Challenge, significantly improving the performance of the same decision model without the proposed semantic ranking.

In order to semantically discern relevant and irrelevant documents for a given question, we are designing a set of discriminators, each receiving a tuple *(question, candidate answer, and document)* and returning a confidence score. A higher score means that the document contains relevant information for answering the question, whereas a score equal to 0 signifies a document that can be ignored as it is noisy or unrelated. These discriminators are the core idea in our approach and they will be used further on to learn how to rank the documents. As will be expanded, these are deep learning models pre-trained to achieve some particular (semantically related) objective. In the next sections, various discriminators are illustrated, and, after that, we present how they are combined to produce the final ranking model.

## 3.3 Document relevance discriminator

This discriminator's purpose is to determine whether or not the document has any significance in answering the question at hand. It ignores the candidate answer and takes into account only the question and the document. The intuition is that some documents are clearly not helpful for the question as they do not contain any relevant information. For example, consider the question: "*How many electrons does a hydrogen atom have?*" and the following two possible extracted documents: "*The hydrogen atom is an electrically neutral atom, usually denoted using the symbol H.*", and "*The hydrogen is a chemical element with a single electron orbiting its nucleus.*" It is clear that the second context is relevant, whilst the first one is not that important. Therefore, in the ranking process, we should place the latter, lower in the list of candidate documents, because semantically it is not relevant for the question at hand, although it has a large TF-IDF score.

| Model | Validation Accuracy |
|---|---|
| GRU with GloVe | 59.04% |
| GRU with ELMo | 60.90% |
| BERT Base | 75.30% |
| BERT Large | 80.43% |

Table 1: Discriminative performance of the models on the adapted SQuAD 2.0 validation set

We have constructed this discriminator by training a neural network on an adapted version of the SQuAD 2.0 dataset (Rajpurkar et al., 2018). The network receives a question and a document as input and produces a score between 0 and 1, correlated with the significance of the document for the question.

The SQuAD 2.0 dataset has the following structure: a paragraph of raw text and a question targeting the information in the paragraph. The answer can be either a span from the paragraph or "*not answerable*", meaning that an answer cannot be deduced based solely on the information in the paragraph. Observe the duality of the task: a "*not answerable*" entry also means that the paragraph is not relevant to the question. This is the key insight that we are going to exploit. We adopted the SQuAD 2.0 dataset, but translated the task into a binary one: is the question answerable or not? Notice that if an answer exists, we pay no attention to the answer itself, the important fact being that the paragraph contains an answer for the question. The resulting dataset has about 100k questions where an answer exists and 50k questions with no valid answer in the given paragraph.

The dataset size is generous enough to enable training deep neural architectures. We have tried four different neural networks to play the role of a document relevance discriminator. The first model encodes the question and the document via 2 layers of independent Gated Recurrent Unit (GRU) cells

(Cho et al., 2014) with the words embedded into a 50-dimensional vector space using pre-trained GloVe word vectors (Pennington et al., 2014). Replacing GRU with Long-Short Term Memory (LSTM) cells (Hochreiter and Schmidhuber, 1997) gives similar performance but the training procedure is computationally more expensive. On top of the representations computed by the GRUs, we added 2 layers of feed-forward connections, the decision phase of the network.

In the second approach, we replaced the first layer of GRUs with the more expressive ELMo encoder (Peters et al., 2018). The other parts of the architecture have not been modified.

Last but not least, we deployed a pre-trained transformer, BERT (Devlin et al., 2018), both the base version with 12 layers and the large version with 24 layers of transformers. BERT is currently the state-of-the-art approach in multiple NLP tasks including open-domain question answering and reading comprehension (Devlin et al., 2018). We fine-tuned the model on the SQuAD 2.0 dataset, with the following hyperparameters: learning rate: 3e-5, warm-up steps: 10% (refer to the BERT paper (Devlin et al., 2018) for further details), sequence length: 425 tokens, batch size: 24 for BERT Large and 10 for BERT base. In order to fine-tune BERT Large, we employed one Tensor Processing Unit (TPU) for about 2 hours. The tokenization has been done following the recommendations in the original paper (Devlin et al., 2018) and the input to the network as the following structure: *"[CLS] question [SEP] supporting document [SEP]"*.

These four different discriminators were trained with the objective of discerning between relevant and irrelevant supporting documents for a given question. Their binary accuracies are reported in Table 1. All models were trained on the modified SQuAD 2.0 train split and the performance is reported on the validation dataset. The effectiveness of the BERT Large model is much better than the others, as expected. Therefore, this fine-tuned model is chosen as the final document relevance discriminator (DRD).

## 3.4 Answer verifier discriminator

The purpose of the second discriminator is to probe whether the answer can be inferred to be correct given the supporting document. Of course, we want to rank documents taking into account if they can be used to find the correct answer. Please notice the core difference between the DRD and the answer verifier discriminator (ARD). The latter is also considering the answer at hand, whereas the former only examines the question and the given supporting document.

Training the discriminator is performed on the RACE (Reading Comprehension Dataset) which is also collected from English Examinations (Lai et al., 2017). The structure of the dataset is perfect for our situation. It contains multiple-choice questions with relevant supporting paragraphs for the correct answer. It is guaranteed that the answer can be deduced by understanding the information in the associated paragraph. We transform this dataset into tuples *(question, candidate answer, paragraph)* that are labeled either as negative, meaning that the answer cannot be verified using the paragraph (for incorrect answers), or as positive – reinforcing the fact that the paragraph can be used to infer that the answer is correct. The RACE dataset is extremely suitable in this situation as it always provides a paragraph that is relevant to the question. This is not the case for the ARC dataset, in which no supporting document is given – thus, one has to be extracted and it is not guaranteed to be always the "right" document.

We have fine-tuned the BERT Large model on the joined RACE middle-school and RACE high-school datasets, totaling near 100k questions with 28k documents. Each question generates 4 entries for the ARD: the 3 wrong answers generate 3 negative examples and the correct answer generates one positive example.

The hyperparameters used for fine-tuning BERT are similar to the DRD discriminator, with differences in the maximum sequence length (512 tokens for RACE vs. 425 tokens for SQuAD 2.0). These hyperparameters were found by trying multiple sensible assignments and selecting the best one according to the validation error. Results are reported in the next section on the test split.

Each tuple *(question, answer, and document)* is fed into the BERT model as: *"[CLS] question [SEP] answer [SEP] document [SEP]"*. The final accuracy of the model on the merged RACE middle and high test datasets is 68.28%.

As a remark, we want to highlight that both the DRD and ARD discriminators have been trained on different datasets than the final multiple choice QA model, which is trained and evaluated on the ARC Easy and Challenge datasets. Hence, our intuition was that transfer learning will succeed even though there are some core differences in the nature of the datasets.

### 3.5 TF-IDF discriminator

We decided to also consider the TF-IDF score computed by Lucene as the third discriminator. This is a lexical (non-semantic) discriminator but it might helpful for simpler questions, especially the ones in the ARC Easy dataset.

### 3.6 Ranking architecture

We have described three discriminators whose purpose is to discern between relevant and useless documents extracted by the IR engine. As mentioned, our proposed approach is to retrieve the first $N$ documents and then rank them taking into account a semantic criterion (as dictated by the pre-trained discriminators). In this section, we are going to describe a neural network that is latently learning to rank the extracted documents whilst selecting the correct answer to the question.

In the ARC dataset, more than 95% of the questions have exactly 4 candidate answers and only one is correct. For a given question and a candidate answer, we pass the extracted $N$ documents though the set of discriminators obtaining a list of scores, one for each discriminator. These scores need to be combined into a final answer score by a mechanism, called Attentive Ranker, inspired from self-attention which we will describe in the next paragraphs.

As shown in Figure 2, each supporting document is associated with a list of scores computed by the discriminators. These scores are projected into a higher dimensional space using a learned projection matrix (we will denote by $D$ the dimension of the projection space – in our experiments, $D = 32$). We then apply an attention mechanism over the semantically labeled documents to select the most relevant ones for a given candidate answer. Let $A$ be a $D \times N$ matrix where each column is the encoding of a document after applying the set of discriminators and the projection step. As we are fetching $N$ documents using the IR engine, matrix $A$ has $N$ columns. Notice that this matrix encodes all the information for a given candidate answer. Next, inspired by the self-attention mechanism (Vaswani et al., 2017), we developed a way to relate documents between each other in order to compute a global representation for them. First of all, we project each row of the matrix $A$ into a key space (with $M$ dimensions) whose purpose is to encode aspects about the quality (relevance) of the associated document (a column in matrix $A$):

$$K = \tanh(W_k A \oplus b_k) \quad (1)$$

where $A \in \mathbb{R}^{D \times N}, W_k \in \mathbb{R}^{M \times D}, b_k \in \mathbb{R}^M$ and by the $\oplus$ operator, we denote the addition of a vector to each column of a matrix (here and throughout the rest of the paper). Second, each column of matrix $A$ is encoded into a value space (with $Q$ dimensions), which is going to dictate the output of the attention mechanism:

$$V = \text{relu}(W_V A \oplus b_V) \quad (2)$$

where $W_V \in \mathbb{R}^{Q \times D}, b_V \in \mathbb{R}^Q$. The intuition is that the value space is dictating what the output of the attention should be, whereas the key space encodes necessary information about which elements (documents in our case) the network should pay attention to. Third, we normalize the key vectors using the softmax function:

$$P = softmax(W_P K \oplus b_P) \quad (3)$$

where $W_P \in \mathbb{R}^{1 \times M}, b_P \in \mathbb{R}$ and the bias is added to each element of $P \in \mathbb{R}^{1 \times N}$. As a result, the weights in $P$ dictate to what extent the network should attend each document. The output of the layer is the weighted sum of the value vectors:

$$Y = VP^T \quad (4)$$

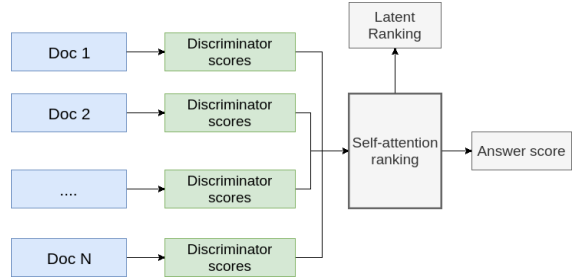

Figure 2: Applying discriminators for a candidate answer and ranking the documents

The same procedure is applied to all candidate answers and the attention weights are shared because we want to have the same representation regardless of the position of the answer in the candidate list (e.g. answer A, B, C, or D). Figure 2 shows the entire encoding procedure and how the attention network is applied.

Observe the fact that for each candidate answer a score (a vector with the dimension of the value space) is obtained by applying a general function. This score is further used to infer which of the four possible answers is, indeed, correct. This decision

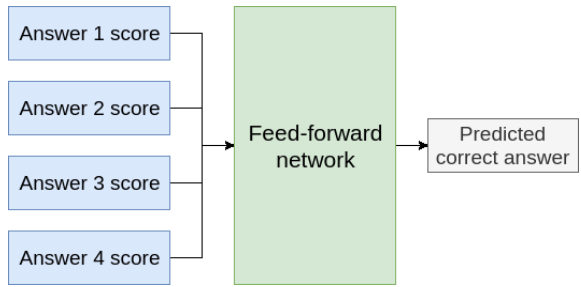

Figure 3: Applying discriminators for a candidate answer and ranking the documents

step is accomplished by a simple feed-forward network (Figure 3).

As a result, one can view this architecture as a feedback loop system. The neural network learns to predict the correct answer. In order to predict the correct answer, it is constrained to select more relevant documents. More relevant documents result in better classification, and so on. Thus, the proposed QA self-attention model, called Attentive Ranker, is jointly *Answering questions by learning to rank* and *Learning to rank by answering questions*.

In the following section, we highlight how the Attentive Ranker can improve QA systems by combining semantic information from a small number of supporting documents extracted with an IR engine.

## 4 Results

We report our model performance on two of the most important multiple-choice science questions datasets: ARC Easy and ARC Challenge (Table 2). During training, due to the small size of the datasets, we experienced a large variance in performance. To partially overcome this issue, we have trained the models a number of times with different random initialization and kept the weights producing the lowest validation loss as the final weights for the QA model.

We have used two knowledge bases in order to retrieve external supporting information: the ARC Corpus (Clark et al., 2018) which contains 14M science-related sentences and a book collection (Pîrtoacă et al., 2018) consisting of about 35 books crawled from online resources such as CK12[3]. Hereinafter, when we mention top *N* retrieved documents, it means that top *N / 2* are fetched from ARC Corpus and *N / 2* are extracted from the book collection, giving the two knowledge bases equal importance.

[3] https://www.ck12.org/

| Dataset | Train | Dev | Test |
|---|---|---|---|
| ARC Easy | 2,251 | 570 | 2,376 |
| ARC Challenge | 1,119 | 299 | 1,172 |

Table 2: Number of questions in the ARC dataset

| Model | Accuracy |
|---|---|
| Random guess | 25.00% |
| IR Solver | 62.55% |
| Reading Strategies (previous SOTA) | 68.90% |
| ***Attentive Ranker (ours)*** | ***72.30%*** |

Table 3: Results on ARC Easy test

| Model | Accuracy |
|---|---|
| Random guess | 25.00% |
| BERT (our implementation) | 40.00% |
| Reading Strategies | 42.32% |
| BERT (previous SOTA - Microsoft) | 44.62% |
| ***Attentive Ranker (ours)*** | ***44.72%*** |

Table 4: Results on ARC Challenge test

| Dataset | TFD | +DRD | +AVD |
|---|---|---|---|
| ARC Easy | 63.89% | 67.48% | 72.30% |
| ARC Challenge | 26.70% | 34.16% | 44.72% |

Table 5: The impact of adding more discriminators on the test set accuracy

The ranking neural network is trained for about 50 epochs, batch size 128, with categorical cross entropy loss, optimized using Adam (Kingma and Ba, 2014).

First, it is important to observe the impact of the number of documents, $N$, on the model's performance (refer to Figure 4 and Figure 5). Considering more than one document improves the accuracy, thus verifying our initial assumption that relevant supporting documents can be found in the first $N$, but it is not always the case that the first one is the most relevant. As it can be observed in Figures 4 and 5, there is a large improvement in accuracy when increasing from 1 to 5 documents. The accuracy continues to increase for both ARC Easy and ARC Challenge until it reaches a maximum at about 40 documents. Thus, for the next experiments, we set the number of documents to 40 (20 from the ARC corpus and 20 from the science book collection).

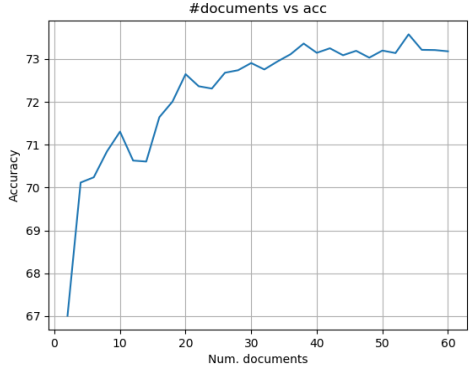

Figure 4: Performance vs. number of documents measured on the ARC Easy validation set

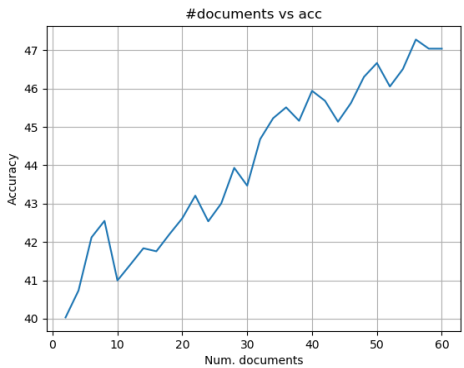

Figure 5: Performance vs. number of documents measured on the ARC Challenge validation set

| Dataset | # docs | Ranking | Accuracy (Δ) |
|---|---|---|---|
| Val. | Top 1 | TF-IDF | 35.59% |
| Val. | Top 1 | Ours | *38.30% (+2.71)* |
| Val. | Top 10 | TF-IDF | 35.93% |
| Val. | Top 10 | Ours | *43.72% (+7.79)* |
| Test | Top 1 | TF-IDF | 34.93% |
| Test | Top 1 | Ours | *37.51% (+3.58)* |
| Test | Top 10 | TF-IDF | 37.08% |
| Test | Top 10 | Ours | *40.00% (+2.92)* |

Table 6: Downstream model performance on the ARC Challenge dataset comparing ranking with Attentive Ranker vs. TF-IDF

| Dataset | # docs | Ranking | Accuracy (Δ) |
|---|---|---|---|
| Val. | Top 1 | doc2vec | 36.61% |
| Val. | Top 1 | Ours | *38.30% (+1.69)* |
| Val. | Top 10 | doc2vec | 39.66% |
| Val. | Top 10 | Ours | *43.72% (+4.06)* |
| Test | Top 1 | doc2vec | 33.90% |
| Test | Top 1 | Ours | *37.51% (+3.61)* |
| Test | Top 10 | dov2vec | 37.85% |
| Test | Top 10 | Ours | *40.00% (+2.15)* |

Table 7: Downstream model performance on the ARC Challenge dataset comparing ranking with Attentive Ranker vs. doc2vec

Second, we verified the model performance on the ARC test sets in order to check how the model generalizes on unseen data and to compare it with other top models in the ARC public leaderboard (https://leaderboard.allenai.org/arc/submissions/public). A summary of the results is reported in Table 3 and Table 4. In both cases, our Attentive Ranker model outperforms the current state-of-the-art (SOTA) approach proving that, indeed, performing a semantic ranking is very effective for QA systems. For a fair comparison, we also reported BERT results as obtained in our implementation and training. The previous SOTA was held by a Microsoft implementation of BERT.

As an ablation study, we wanted to identify the contribution of the discriminators to the overall performance of the Attentive Ranker. We performed an experiment, where starting from the TF-IDF lexical discriminator, other discriminators are incrementally added to the scheme: document relevance discriminator (DRD) and answer verifier discriminator (AVD). The results are revealed in Table 5 where we measured the total number of questions correctly answered in the ARC Easy and ARC Challenge test sets. The influence of both semantic-level discriminators (DRD, AVD) is fundamental: they increase the performance with up to 20%, both contributing to the end-to-end accuracy. Notice that the discriminators have a greater impact on the challenge questions as those are radically more difficult and require high-level reasoning to determine the correct answer (that is why the TF-IDF discriminator, which works at a lexical level, has a performance close to random).

One other examination that we enforced is to analyze whether the ranked documents that we computed (semantically sorted by our Attentive Ranker) are more relevant than the documents sorted by the TF-IDF metric. We fine-tuned a downstream BERT (Devlin et al., 2018) model that selects the most probable answer from the candidate list given a list of supporting documents. In the experiment split A, we fed this model with top *N* documents as ranked by the TF-IDF metric and in experiment split B we fed the model with top *N* documents as sorted by our ranking neural network (given the weights in the attention layer).

As we have previously shown, the ARC Challenge dataset is more eloquent for this type of semantical analysis due to the difficult questions it includes, thus, we use it for this investigation.

Table 6 shows that using documents ranked by our attentive neural network always leads to a performance increase in downstream models, compared to TF-IDF. On the validation set, the improvement is considerably higher (+7.79) due to a possible over-fitting of the hyperparameters during the Attentive Ranker's training.

We also investigated the ability of the Attentive Ranker to sort the retrieved documents as compared to document embeddings – which may be used to measure the similarity between a question and a candidate document. The question and the documents are embedded using doc2vec (Le and Mikolov, 2014) pre-trained on English Wikipedia. Then, the documents are sorted by their cosine distance to the question. Therefore, the documents that are cosine-closest to a question should be more relevant in answering that question as they are more similar to the question in the doc2vec embeddings space. We would like to note that this approach, although reasonable, may suffer from a subtle problem: the question and the documents have different structures and it may be difficult for an embedding function to capture similarities defined as relevance: "the document is helpful in answering the question". Table 7 illustrates the performance of a downstream model (BERT) in answering the questions when fed with top $N$ (1 or 10) documents as ranked by Attentive Ranker or doc2vec. In all cases, the Attentive Ranker provided more relevant documents than doc2vec, to the BERT-based decision layer. On another hand, doc2vec yields better accuracies than the simple TF-IDF metric.

Our hypothesis that the Attentive Ranker is suitable to rank the retrieved documents by their semantic value to the question is confirmed. Therefore, we decided to make the ranked set of documents public[4], in order to be used by other models. Please refer to the GitHub repository for instructions on how to use the ranked documents. In the same repository, one can find the source code and the full set of trained models.

## 5 Conclusion

In this paper, we highlighted an important problem with many of the current approaches developed for multiple-choice question answering tasks. To overcome the poor performance of the IR engines used to retrieve supporting documents for *(question, candidate answer)* pairs, we described a method that semantically ranks the extracted supporting documents. For this, we proposed an attention-based neural network that latently learns to rank supporting documents by their relevance in answering the given question. The Attentive Ranker architecture depends on the existence of semantic discriminators which are pre-trained to distinguish between relevant and pointless documents. Our proposed model achieves state-of-the-art accuracy on two significant datasets: ARC Easy (72.30%) and ARC Challenge (44.72%).

Furthermore, we have shown that just replacing TF-IDF sorted documents with documents provided by our enhanced ranking method, highly improves the accuracy of various downstream decision models, by up to 7% in our experiments. Therefore, we have made the ranked documents public and further research can benefit from it.

## References

Białecki, A., Muir, R., Ingersoll, G., & Imagination, L. (2012, August). *Apache lucene 4*. In SIGIR 2012 workshop on open source information retrieval (p. 17).

Banerjee, P., Pal, K. K., Mitra, A., & Baral, C. (2019). Careful selection of knowledge to solve open book question answering. arXiv preprint arXiv:1907.10738.

Brysbaert, M., Warriner, A.B., & Kuperman, V. (2014). *Concreteness ratings for 40 thousand generally known English word lemmas*. Behavior research methods, 46 3, 904-11.

Chen, D., Fisch, A., Weston, J., & Bordes, A. (2017). Reading wikipedia to answer open-domain questions. arXiv preprint arXiv:1704.00051.

Cho, K., Van Merriënboer, B., Gulcehre, C., Bahdanau, D., Bougares, F., Schwenk, H., & Bengio, Y. (2014). Learning phrase representations using RNN encoder-decoder for statistical machine translation. arXiv preprint arXiv:1406.1078.

Clark, P., Cowhey, I., Etzioni, O., Khot, T., Sabharwal, A., Schoenick, C., & Tafjord, O. (2018). Think you

---

[4] https://bit.ly/2ZnBNLs

have solved question answering? try arc, the ai2 reasoning challenge. arXiv preprint arXiv:1803.05457.

Clark, P., Etzioni, O., Khot, T., Sabharwal, A., Tafjord, O., Turney, P.D., & Khashabi, D. (2016). *Combining Retrieval, Statistics, and Inference to Answer Elementary Science Questions*. AAAI.

Devlin, J., Chang, M. W., Lee, K., & Toutanova, K. (2018). Bert: Pre-training of deep bidirectional transformers for language understanding. arXiv preprint arXiv:1810.04805.

Gormley, C., & Tong, Z. (2015). *Elasticsearch: The definitive guide: A distributed real-time search and analytics engine.* O'Reilly Media, Inc.

Hochreiter, S., & Schmidhuber, J. (1997). *Long short-term memory*. Neural computation, 9(8), 1735-1780.

Jansen, P., Sharp, R., Surdeanu, M., & Clark, P. (2017). Framing qa as building and ranking intersentence answer justifications. Computational Linguistics, 43(2), 407-449. https://www.doi.org/10.1162/COLI_a_00287

Khashabi, D., Khot, T., Sabharwal, A., & Roth, D. (2017). Learning What is Essential in Questions. CoNLL.

Kingma, D. P., & Ba, J. (2014). Adam: A method for stochastic optimization. arXiv preprint arXiv:1412.6980.

Lai, G., Xie, Q., Liu, H., Yang, Y., & Hovy, E. (2017). Race: Large-scale reading comprehension dataset from examinations. arXiv preprint arXiv:1704.04683.

Le, Q., & Mikolov, T. (2014, January). *Distributed representations of sentences and documents*. In International conference on machine learning (pp. 1188-1196).

Manning, C., Raghavan, P., & Schütze, H. (2010). *Introduction to information retrieval. Natural Language Engineering*, 16(1), 100-103.

Ni, J., Zhu, C., Chen, W., & McAuley, J. (2018). Learning to attend on essential terms: An enhanced retriever-reader model for scientific question answering. arXiv preprint arXiv:1808.09492.

Nicula, B., Ruseti, S., & Rebedea, T. (2018, March). *Improving Deep Learning for Multiple Choice Question Answering with Candidate Contexts*. In European Conference on Information Retrieval (pp. 678-683). Springer, Cham.

Pan, X., Sun, K., Yu, D., Ji, H., & Yu, D. (2019). Improving Question Answering with External Knowledge. arXiv preprint arXiv:1902.00993.

Pennington, J., Socher, R., & Manning, C. (2014). Glove: Global vectors for word representation. In Proceedings of the 2014 conference on empirical methods in natural language processing (EMNLP) (pp. 1532-1543).

Peters, M. E., Neumann, M., Iyyer, M., Gardner, M., Clark, C., Lee, K., & Zettlemoyer, L. (2018). Deep contextualized word representations. arXiv preprint arXiv:1802.05365.

Pirtoaca, G., Rebedea, T., & Ruseti, S. (2018). Improving Retrieval-Based Question Answering with Deep Inference Models. 2019 International Joint Conference on Neural Networks (IJCNN), 1-8.

Pirtoaca, G. S., Ruseti, S., & Rebedea, T. (2018). Improving multi-choice question answering by identifying essential terms in questions. Romanian Journal of Human-Computer Interaction, 11(2), 145-162.

Radford, A. (2018). *Improving Language Understanding by Generative Pre-Training*.

Radford, A., Wu, J., Child, R., Luan, D., Amodei, D., & Sutskever, I. (2019). *Language models are unsupervised multitask learners*. OpenAI Blog, 1, 8.

Rajpurkar, P., Zhang, J., Lopyrev, K., & Liang, P. (2016). Squad: 100,000+ questions for machine comprehension of text. arXiv preprint arXiv:1606.05250.

Rajpurkar, P., Jia, R., & Liang, P. (2018). Know What You Don't Know: Unanswerable Questions for SQuAD. arXiv preprint arXiv:1806.03822. translation. arXiv preprint arXiv:1406.1078.

Sun, K., Yu, D., Yu, D., & Cardie, C. (2018). Improving machine reading comprehension with general reading strategies. arXiv preprint arXiv:1810.13441.

Vaswani, A., Shazeer, N., Parmar, N., Uszkoreit, J., Jones, L., Gomez, A. N., ... & Polosukhin, I. (2017). *Attention is all you need*. In Advances in neural information processing systems (pp. 5998-6008).

Zhang, Y., Dai, H., Toraman, K., & Song, L. (2018). KG^ 2: Learning to Reason Science Exam Questions with Contextual Knowledge Graph Embeddings. arXiv preprint arXiv:1805.12393.